\documentclass[11pt]{article}
\usepackage{coling2020}
\usepackage{times}
\usepackage{url}
\usepackage{latexsym}

\usepackage{multirow}
\usepackage{float}
\usepackage{graphics}
\usepackage{longtable}

\usepackage{hyperref}

\usepackage[T5]{fontenc}
\usepackage[utf8]{inputenc}
\usepackage[vietnamese,english]{babel}

\usepackage{fancyhdr}
\usepackage{lastpage}
\usepackage{subfig}

\usepackage{xcolor}
\usepackage{graphicx}
\usepackage{subfig}
\usepackage{appendix}
\usepackage{multirow}
\usepackage{multicol}
\usepackage{longtable}

\usepackage{tabularx}
\usepackage{supertabular}
\usepackage{adjustbox}
\usepackage{pgfplots}
\usepackage{float}
\usepackage{tikz}

\usetikzlibrary{patterns}

\usepackage{booktabs}
\usepackage{amssymb,amsmath,nccmath}
\usepackage{cclicenses}
\usepackage{makecell}
\usepackage{lscape,array}
\newcolumntype{C}[1]{>{\centering\arraybackslash}p{#1}} 
\usepackage[thin, , thinc]{esdiff}

\usepackage{framed}  

\pgfplotsset{
  compat=1.3,
  tick label style={font=\tiny},
  label style={font=\tiny}}

\usetikzlibrary{arrows}
\definecolor{rosso}{RGB}{220,57,18}
\definecolor{giallo}{RGB}{255,153,0}
\definecolor{blu}{RGB}{102,140,217}
\definecolor{verde}{RGB}{16,150,24}
\definecolor{viola}{RGB}{153,0,153}

\usepackage{pgfplots}

\usetikzlibrary{arrows}
\definecolor{rosso}{RGB}{220,57,18}
\definecolor{giallo}{RGB}{255,153,0}
\definecolor{blu}{RGB}{102,140,217}
\definecolor{verde}{RGB}{16,150,24}
\definecolor{viola}{RGB}{153,0,153}

\tikzset{
  chart/.style={
    legend label/.style={font={\scriptsize},anchor=west,align=left},
    legend box/.style={rectangle, draw, minimum size=5pt},
    axis/.style={black,semithick,->},
    axis label/.style={anchor=east,font={\tiny}},
  },
  pie chart/.style={
    chart,
    slice/.style={line cap=round, line join=round,draw=white},
    pie title/.style={font={\bfseries}},
    slice type/.style 2 args={
        ##1/.style={fill=##2},
        values of ##1/.style={}
    }
  }
}

\pgfdeclarelayer{background}
\pgfdeclarelayer{foreground}
\pgfsetlayers{background,main,foreground}

\newcommand{\pie}[3][]{
    \begin{scope}[#1]
    \pgfmathsetmacro{\curA}{90}
    \pgfmathsetmacro{\radius}{1}
    \def\Centre{(0,0)}
    \node[pie title] at (90:1.3) {#2};
    \foreach \v/\s in{#3}{
        \pgfmathsetmacro{\deltaA}{\v/100*360}
        \pgfmathsetmacro{\nextA}{\curA + \deltaA}
        \pgfmathsetmacro{\midA}{(\curA+\nextA)/2}

        \path[slice,\s] \Centre
            -- +(\curA:\radius)
            arc (\curA:\nextA:\radius)
            -- cycle;

  \pgfmathsetmacro{\ysign}{ifthenelse(mod(\midA,360)<=180,3,-1.2)}
  \pgfmathsetmacro{\xsign}{ifthenelse(mod(\midA-90,360)<=180,-1,1.2)}

  \begin{pgfonlayer}{foreground}
        \draw[*-,thin] \Centre ++(\midA:\radius/2) -- 
                              ++(\xsign*0.07*\radius,\ysign*0.2*\radius) -- 
                              ++(\xsign*\radius,0) 
                      node[above,near end,pie values,values of \s]{$\v\%$};
  \end{pgfonlayer}
        \global\let\curA\nextA
    }
    \end{scope}
}

\definecolor{islamicgreen}{rgb}{0.0, 0.56, 0.0}
\definecolor{neongreen}{rgb}{0.22, 0.88, 0.08}

\newcommand{\slice}[4]{
  \pgfmathparse{0.5*#1+0.5*#2}
  \let\midangle\pgfmathresult

  \draw[thick,fill=black!10] (0,0) -- (#1:1) arc (#1:#2:1) -- cycle;

  \node[label=\midangle:#4] at (\midangle:1) {};

  \pgfmathparse{min((#2-#1-10)/110*(-0.3),0)}
  \let\temp\pgfmathresult
  \pgfmathparse{max(\temp,-0.5) + 0.8}
  \let\innerpos\pgfmathresult
  \node at (\midangle:\innerpos) {#3};
}

\colingfinalcopy 

\title{A Vietnamese Dataset for Evaluating Machine Reading Comprehension}

\author{Kiet Van Nguyen$^{1,2}$, Duc-Vu Nguyen$^{1,2}$, Anh Gia-Tuan Nguyen$^{1,2}$, Ngan Luu-Thuy Nguyen$^{1,2}$\\
  $^{1}$University of  Information Technology, Ho Chi Minh City, Vietnam \\
  $^{2}$Vietnam National University, Ho Chi Minh City, Vietnam \\
  {\tt \{kietnv, vund, anhngt, ngannlt\}@uit.edu.vn} }

\date{}

\begin{document}
\maketitle
\begin{abstract}

Over 97 million people speak Vietnamese as their native language in the world. However, there are few research studies on machine reading comprehension (MRC) for Vietnamese, the task of understanding a text and answering questions related to it. Due to the lack of benchmark datasets for Vietnamese, we present the {\bf Vi}etnamese {\bf Qu}estion {\bf A}nswering {\bf D}ataset ({\bf UIT-ViQuAD}), a new dataset for the low-resource language as Vietnamese to evaluate MRC models. This dataset comprises over 23,000 human-generated question-answer pairs based on 5,109 passages of 174 Vietnamese articles from Wikipedia. In particular, we propose a new process of dataset creation for Vietnamese MRC. Our in-depth analyses illustrate that our dataset requires abilities beyond simple reasoning like word matching and demands single-sentence and multiple-sentence inferences. Besides, we conduct experiments on state-of-the-art MRC methods for English and Chinese as the first experimental models on UIT-ViQuAD. We also estimate human performance on the dataset and compare it to the experimental results of powerful machine learning models. As a result, the substantial differences between human performance and the best model performance on the dataset indicate that improvements can be made on UIT-ViQuAD in future research. Our dataset is freely available on our website\footnote{https://sites.google.com/uit.edu.vn/uit-nlp/datasets-projects} to encourage the research community to overcome challenges in Vietnamese MRC.
\end{abstract}

\section{Introduction}
Machine reading comprehension (MRC) is an understanding natural language task that requires computers to understand a text and then answer questions related to it. MRC is an essential core for a range of natural language processing applications such as search engines and intelligent agents (Alexa, Google Assistant, Siri, and Cortana) In order to evaluate MRC models, gold standard resources with question-answer pairs based on documents have to be collected or created by human. Building a benchmark dataset plays a vital role in evaluating natural language processing models, especially for a low-resource language like Vietnamese.
\blfootnote{This work is licensed under a Creative Commons Attribution 4.0 International License. License details: http://creativecommons.org/licenses/by/4.0/}

Typical gold standard MRC resources for English are span-extraction MRC datasets \cite{Raj:16,Raj:18,Tri:16}, cloze-style MRC datasets \cite{Her:15,Hil:15,Cui:16}, multiple-choice MRC datasets \cite{Ric:13,Lai:17} and conversation-based MRC datasets \cite{Red:19,Sun:19}. For other languages, there are the Chinese dataset of the span-extraction MRC \cite{Cui:18,duan2019cjrc}, the traditional Chinese dataset of MRC \cite{Shao:18}, the Chinese user-query-log-based dataset of DuReader \cite{He:17}, and the Korean MRC dataset \cite{Lim:19}. 
Due to the rapid development of the reading comprehension datasets, various neural network-based models have been proposed and made a significant advancement in this research field such as Match-LSTM \cite{Wan:16}, BiDAF \cite{Seo:16}, R-Net \cite{Wang:17}, DrQA \cite{Che:2017}, FusionNet \cite{Hua:17}, FastQA \cite{Wei:17}, and QANet \cite{Yu:18}. Pre-trained language models, BERT \cite{Dev:18} and XLM-R \cite{Con:19} have recently become extremely popular and achieved state-of-the-art performances for MRC tasks.

Vietnamese is a language with few resources for natural language processing. The dataset for MRC introduced by \cite{Ngu:20} consists of 2,783 multiple-choice questions and answers based on a set of 417 Vietnamese texts which are used for evaluating the reading comprehension skill for $1^{st}$ to $5^{th}$ graders. However, this dataset is relatively small in size to evaluate deep learning models for the Vietnamese MRC. Thus, we aim to build a new large dataset for evaluating Vietnamese MRC.

Though the deep learning approach has surpassed the human performance on the SQuAD \cite{Raj:16} and NewsQA \cite{Tri:16} datasets, we wonder if these state-of-the-art models could also achieve similar performances on datasets of different languages. To further enhance the development of the MRC, we build a span-extraction MRC dataset where answers to questions are always spans from a given text for Vietnamese. Figure \ref{fig:examples} shows several examples for Vietnamese span-extraction reading comprehension. In this study, we have four main contributions described as follows.
\begin{itemize}
    \item We create a benchmark dataset for evaluating Vietnamese MRC: UIT-ViQuAD comprises 23,074 human-generated question--answer pairs based on 5,109 passages of 174 Vietnamese Wikipedia articles. The dataset is available freely on our website\footnote{https://sites.google.com/uit.edu.vn/uit-nlp/datasets-projects} for research purposes.
    \item To gain thorough insights into the dataset, we analyze the dataset according to different linguistic aspects including length-based analysis (question length, answer length, and passage length) and type-based analysis (question type, answer type, and reasoning type).
    \item To achieve first MRC evaluation on UIT-ViQuAD, we conduct experiments with MRC models which are state-of-the-art for English and Chinese. Then, we compare performances between the machine models and humans in terms of different linguistic aspects. These in-depth analyses provide insights into span-based MRC in Vietnamese.
    \item Cross-lingual MRC \cite{cui2019cross} is a new trend in natural language processing. Our proposed MRC dataset for Vietnamese could also be a resource for cross-lingual study along with other similar datasets such as SQuAD, CMRC, and KorQuAD.
\end{itemize}

\begin{table}[H]
\resizebox{\columnwidth}{!}{\begin{tabular}{l}
\hline
\begin{tabular}[c]{p{18cm}}\underline{{\bf Passage}}: Nước biển có độ mặn không đồng đều trên toàn thế giới mặc dù phần lớn có độ mặn nằm trong khoảng từ {\it{\bf \textcolor{red}{3,1\%}}} tới 3,8\%. Khi sự pha trộn với nước ngọt đổ ra từ các con sông hay gần các sông băng đang tan chảy thì nước biển nhạt hơn một cách đáng kể. Nước biển nhạt nhất có tại {\it{\bf \textcolor{blue}{vịnh Phần Lan}}}, một phần của biển Baltic.\\ ({\bf English}: Seawater has uneven salinity throughout the world although most salinity ranges from {\bf \textcolor{red}{ 3.1\%}} to 3.8\%. \\ When the mix with freshwater pouring from rivers or near glaciers is melting, the seawater is significantly lighter. The lightest seawater is found in the {\it{\bf \textcolor{blue}{ Gulf of Finland}}}, a part of the Baltic Sea.)\end{tabular} \\ \hline
\begin{tabular}[c]{p{18cm}}\underline{{\bf Question}}: Độ mặn thấp nhất của nước biển là bao nhiêu? ({\bf English}: What is the lowest salinity of seawater?)\\ \underline{{\bf Answer}}: {\it{\bf \textcolor{blue}{3.1\%}}} ({\bf English}: 3.1\%)\end{tabular}                                                                                                                                                                                                                                                                                                                                                                                                                                                                                                                                                                                                                                                                                                           \\ \hline
\begin{tabular}[c]{p{18cm}}\underline{{\bf Question}}: Nước biển ở đâu có hàm lượng muối thấp nhất? ({\bf English}: Where is the lowest salt content?)\\ \underline{{\bf Answer}}: {\it{\bf \textcolor{blue}{Vịnh Phần Lan}}}. ({\bf English}: Gulf of Finland.)\end{tabular}                                                                                                                                                                                                                                                                                                                                                                                                                                                                                                                                                                                                                                                       \\ \hline
\end{tabular}}
\captionof{figure}{Several examples for Vietnamese span-extraction reading comprehension. The English translations are also provided for comparison.}\label{fig:examples}
\end{table}

The rest of this paper is structured as follows. Section 2 reviews existing datasets. Section 3 introduces the creation process of our dataset. In-depth analyses of our dataset are presented in Section 4. Then Section 5 presents our experiments and analysis results. Finally, Section 6 presents conclusions and directions for future work.

\section{Existing datasets}
Because we aim to build a span-based MRC dataset for Vietnamese, a range of recent span-extraction MRC datasets such as SQuAD \cite{Raj:16}, NewsQA \cite{Tri:16}, CMRC \cite{Cui:18}, and KorQuAD \cite{Lim:19} is reviewed in this section. These datasets are described as follows.

\textbf{SQuAD} is one of the most popular English datasets of the span-based MRC. Rajpurkar et al. ~\shortcite{Raj:16}. proposed SQuAD v1.1 created by crowd-workers on 536 Wikipedia articles with 107,785 question-answer pairs. SQuAD v2.0 \cite{Raj:18} was released with adding over 50,000 unanswerable questions created adversarially by crowd-workers according to the original ones.

\textbf{NewsQA} is another English dataset proposed by Trischler et al. \shortcite{Tri:16}, consisting of 119,633 question-answer pairs generated by crowd-workers on 12,744 news articles from CNN news. This dataset is similar to SQuAD because the answer to each question is a text segment of arbitrary length in the corresponding news article. 

\textbf{CMRC} \cite{Cui:18} is a span-extraction dataset for Chinese MRC introduced in the Second Evaluation Workshop on Chinese Machine Reading Comprehension 2018, comprising approximately 20,000 human-annotated questions on Wikipedia articles.

\textbf{KorQuAD} \cite{Lim:19} is a Korean dataset for span-based MRC, consisting of over 70,000 human-generated question-answer pairs on Korean Wikipedia articles.

These datasets are studied in the development and evaluation of various deep neural network models in NLP, such as Match-LSTM \cite{Wan:16}, BiDAF \cite{Seo:16}, R-Net \cite{Wang:17}, DrQA \cite{Che:2017}, FusionNet \cite{Hua:17}, FastQA \cite{Wei:17} and QANet \cite{Yu:18}. Most recently, BERT \cite{Dev:18} and XLM-R \cite{Con:19}, which are powerful models trained on multiple languages, have obtained state-of-the-art performances on MRC datasets.

Until now, there has not been any datasets of Vietnamese Wikipedia texts for span-based MRC research. As mentioned above, the datasets are benchmarks for the MRC task and may be used for organizing a challenge which encourages researchers to explore the best processing models. Therefore, this is our primary motivation to create the new dataset for Vietnamese MRC.





\section{Dataset creation}
\label{datacreation}

In this section, we introduce our proposed process of MRC dataset creation for the Vietnamese language. In particular, we build our UIT-ViQuAD dataset through five phases consisting of worker recruitment, passage collection, question-answer sourcing, validation and additional answers collection. These phases are described in detail as follows.

\label{dataset}
\begin{figure}[H]
  \centering
  \includegraphics[width=0.8\linewidth]{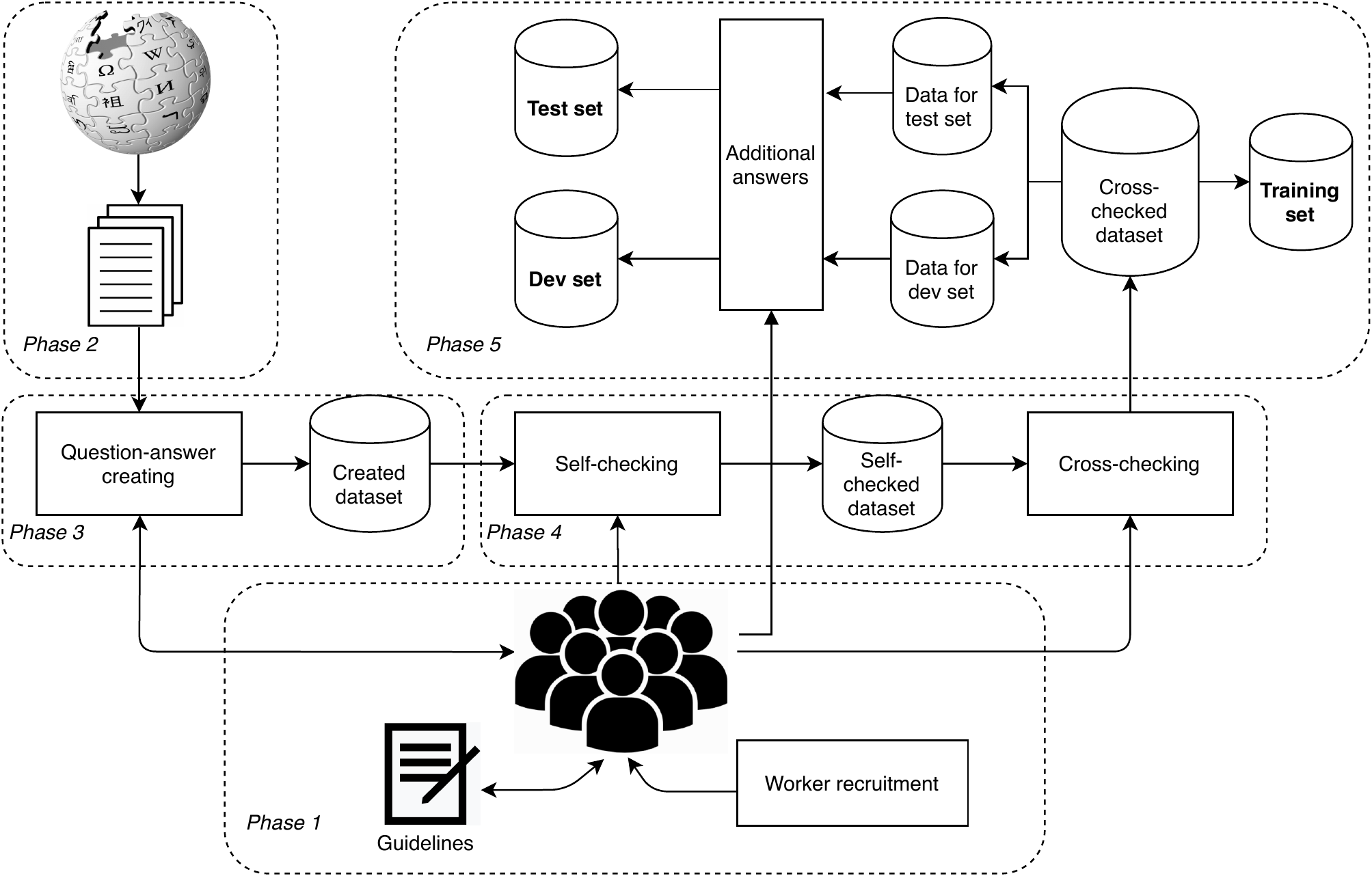}
  \caption{The overview process of creating our dataset UIT-ViQuAD.}
  \label{fig:datacreationprocess}
\end{figure}


{\bf Phase 1 - Worker recruitment}: The quality of a dataset depends on high-quality workers and the process of data creation. In this section, we present worker recruitment for creating our dataset according to a rigorous process, consisting of four different stages. (1) People apply to become workers for creating answer–question pairs of the dataset; (2) Selected people are excellent at general knowledge and passed our reading comprehension test; (3) Official workers are carefully trained over 500 question-answer pairs and cross-checked their created data to detect common mistakes that can be avoided when creating data. 

{\bf Phase 2 - Passage collection}: Similar to SQuAD, we also use Project Nayuki's Wikipedia's internal PageRanks\footnote[3]{\url{https://www.nayuki.io/page/computing-wikipedias-internal-pageranks}} to obtain a set of the top 5,000 Vietnamese articles, from which we choose randomly 151 articles for dataset creation. Each passage corresponds to a paragraph in an  article. Images, figures, and tables are excluded. We also delete passages shorter than 300 characters or containing many special characters and symbols.

{\bf Phase 3 - Question-answer sourcing}: Workers comprehend each passage and then create questions and corresponding answers. During the question and answer creation, workers follow rules which are: (1) Workers are required to create at least three questions per passage. (2) Workers are encouraged to ask questions in their own words. (3) Answers are text spans in the passage that are used to answer the questions. (4) Workers are encouraged to make diversities in questions, answers, and reasoning.

{\bf  Phase 4 - Question and answer validation}: In this phase, workers perform two different sub-phases to check mistakes in question-answer pairs including self-checking and cross-checking. The mistakes are classified into five different categories: unclear questions, misspellings, incorrect answers, lack or excess of information in answers, and incorrect-boundary answers. The two sub-phases are described as follows.
\begin{itemize}
\item \textbf{Self-checking}: Workers revise their question-answer pairs themselves. 
\item \textbf{Cross-checking}: Workers cross-check each other's question-answer pairs. If they discover any mistakes in the dataset, they discuss with each other to correct the  mistakes.
\end{itemize}

{\bf Phase 5 - Additional answers collection}: To evaluate the quality of dataset creation, for the development and test datasets, we add three more answers for each question by different workers in addition to the original answer. During this phase, the workers cannot see each other's answer and they are encouraged to make diversified answers. 

\section{Dataset analysis}
\label{datasetanalysis}

\subsection{Overall statistics}
The statistics of the training (Train), development (Dev) and test (Test) sets of our dataset are described in Table \ref{tab:overallstatistics}. The number of questions of UIT-ViQuAD is 23,074. In the table, the numbers of articles and passages, the average lengths\footnote[6]{We use the pyvi library \url{https://pypi.org/project/pyvi/} for word segmentation.} of questions and answers, and vocabulary sizes are also presented.

\begin{table}[H]
\centering
\begin{tabular}{lrrrr}
\hline
\multicolumn{1}{c}{\textbf{}} & \multicolumn{1}{c}{\textbf{Train}} & \multicolumn{1}{c}{\textbf{Dev}} & \multicolumn{1}{c}{\textbf{Test}} & \multicolumn{1}{c}{\textbf{All}} \\ \hline
Number of articles              & 138                                        & 18                                            & 18                                     & 174                               \\ 
Number of passages              & 4,101                                      & 515                                           & 493                                    & 5,109                             \\ 
Number of questions             & 18,579                                     & 2,285                                         & 2,210                                  & {\bf 23,074}                            \\ 
Average passage length          &153.9                                            &147.9                                               &155.0                                        &153.4                                   \\ 
Average question length         & 12.2                                       & 11.9                                          & 12.2                                   & 12.2                              \\ 
Average answer length           & 8.1                                        & 8.4                                           & 8.9                                    & 8.2                               \\ 
Vocabulary size                 & 36,174                                     & 9,184                                         & 9,792                                  & 41,773                            \\ \hline
\end{tabular}
\caption{Overview statistics of the UIT-ViQuAD dataset.}
\label{tab:overallstatistics}
\end{table}

\subsection{Length-based analysis}
We present statistics of our dataset according to three types of length including question length (see Table \ref{tab:qalength}), answer length (see Table \ref{tab:qalength}), and passage length (see Table \ref{tab:readingtextlength}). The 11-15-word questions of the dataset account for a high proportion of 45.29\%. The answers are mostly from 1 to 10 word lengths, accounting for 73.68\%. The length of passages is largely from 101 to 200 words with 73.13\%. These analyses show that our dataset has its own characteristics.  


\begin{table}[!htb]
\resizebox{8cm}{!}{
\begin{minipage}[b]{.65\textwidth}
    \centering

\begin{tabular}{crrrr|rrrr}
\hline
\multirow{2}{*}{\textbf{Length}} & \multicolumn{4}{c}{\textbf{Question}}                                                                                                           & \multicolumn{4}{c}{\textbf{Answer}}                                                                                                             \\ \cline{2-9} 
                                 & \multicolumn{1}{c}{\textbf{Train}} & \multicolumn{1}{c}{\textbf{Dev}} & \multicolumn{1}{c}{\textbf{Test}} & \multicolumn{1}{c}{\textbf{All}} & \multicolumn{1}{c}{\textbf{Train}} & \multicolumn{1}{c}{\textbf{Dev}} & \multicolumn{1}{c}{\textbf{Test}} & \multicolumn{1}{c}{\textbf{All}} \\ \hline
1-5                              & 1.03                                & 1.44                              & 0.95                               & 1.06                              & {\bf 54.12}                               & {\bf 50.63}                             & {\bf 52.26}                              & {\bf 53.60}                             \\ 
6-10                             & 35.99                                & 38.38                             & 33.21                              & 35.96                             & 19.95                               & 22.14                             & 19.10                              & 20.08                             \\ 
11-15                            & {\bf 44.97}                               & {\bf 44.29}                             & {\bf 49.05}                              & {\bf 45.29}                             & 10.86                               & 10.81                             & 10.81                              & 10.85                             \\ 
16-20                            & 15.01                               & 13.61                             & 14.07                              & 14.78                             & 6.28                                & 7.48                              & 6.83                               & 6.45                              \\ 
\textgreater{}20                 & 3.00                                & 2.28                              & 2.71                               & 2.90                              & 8.80                                & 8.93                              & 11.00                              & 9.02                              \\ \hline
\end{tabular}
\caption{Statistics of the question and answer lengths on our dataset.}
    \label{tab:qalength}
\end{minipage}}\hfill
\resizebox{7.5cm}{!}{\begin{minipage}[b]{.7\textwidth}
    \centering

\begin{tabular}{crrrr}
\hline
\multirow{2}{*}{\textbf{Length}} & \multicolumn{4}{c}{\textbf{Passage}}                                                                                                            \\ \cline{2-5} 
                                 & \multicolumn{1}{c}{\textbf{Train}} & \multicolumn{1}{c}{\textbf{Dev}} & \multicolumn{1}{c}{\textbf{Test}} & \multicolumn{1}{c}{\textbf{All}} \\ \hline
\textless{}101                   & 11.41                               & 10.10                             & 11.16                              & 11.25                             \\ 
101-150                          & {\bf 47.50}                               & {\bf 53.59}                             & {\bf 45.44}                              & {\bf 47.92}                             \\ 
151-200                          & 24.99                               & 23.69                             & 28.60                              & 25.21                             \\ 
201-250                          & 9.41                                & 8.93                              & 9.94                               & 9.41                              \\ 
251-300                          & 4.02                                & 2.52                              & 1.83                               & 3.66                              \\ 
\textgreater{}300                & 2.66                                & 1.17                              & 3.04                               & 2.54                              \\ \hline
\end{tabular}
\caption{Statistics of the passage lengths on our dataset.}
    \label{tab:readingtextlength}
\end{minipage}}
\end{table}

\subsection{Type-based analysis}
\begin{figure}[H]
    \centering
    \subfloat[Question Type]{{\includegraphics[width=6cm]{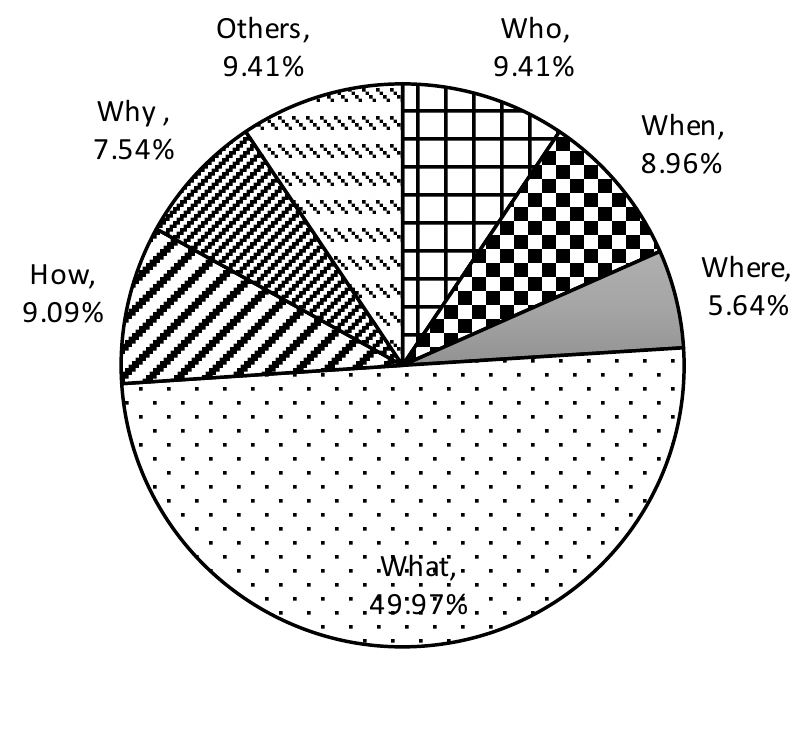} }\label{fig:questiontype}}%
    \qquad
    \subfloat[Reasoning Type]{{\includegraphics[width=6cm]{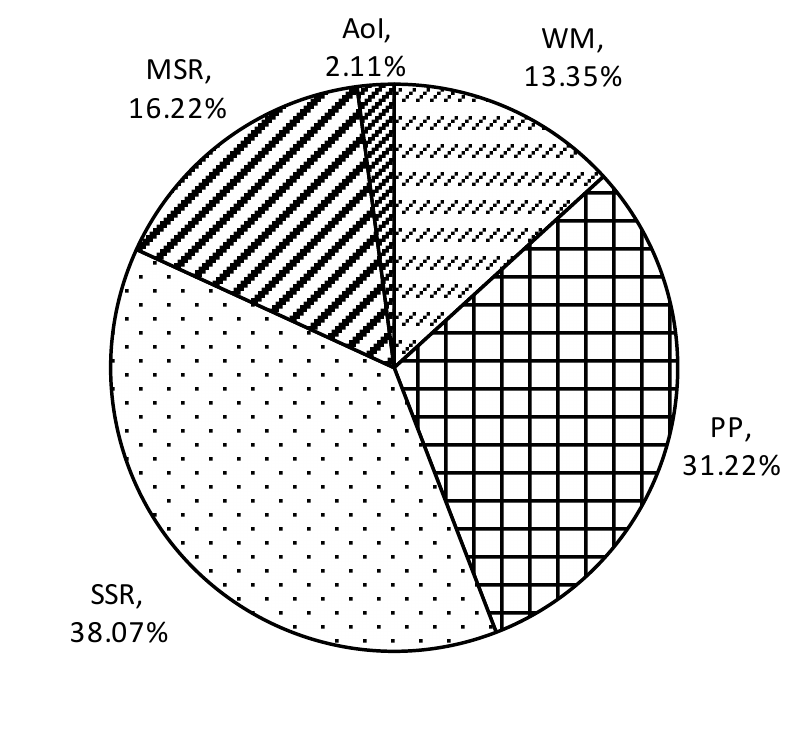} }\label{fig:reasoningtype}}%
    \caption{The distribution of the question and reasoning types on the Dev set of UIT-ViQuAD.}%
    \label{fig:example}%
\end{figure}

In this section, we analyze the Dev set in terms of different types such as {\it question type}, {\it reasoning type}, and {\it answer type}. Because Vietnamese is a subject-verb-object language similar to Chinese \cite{nguyen2018ensuring}, Vietnamese question types in UIT-ViQuAD follow a manner in CMRC \cite{Cui:18}. Thus, we also divide the questions into seven types: Who, What, When, Where, Why, How, and Others. However, in Vietnamese, question words vary a lot, so we have Workers manually annotate the type of questions. Figure \ref{fig:questiontype} presents the distribution of the question types on our dataset. What questions account for the largest proportion of 49.97\%. Compared to SQuAD, the percentage of the What question in our dataset is similar to that in SQuAD (53.60\%) \cite{aniol2019ensemble}.

To explore the difficulty of reasoning required, we conduct human annotation for the different reasoning level of the question, shown in Figure \ref{fig:reasoningtype}. Following Hill et al. \shortcite{Hil:15} and Nguyen et al. \shortcite{Ngu:20}, workers manually annotate the questions into five different types of reasoning with ascending order of difficulty: 
word matching (WM), paraphrasing (PP), single-sentence reasoning (SSR), multi-sentence reasoning (MSR), and ambiguous/insufficient (AoI). Our dataset is more difficult than SQuAD and NewsQA because the percentage of inference types (68.29\%) in our dataset is higher than that in SQuAD (20.5\%) and NewsQA (33.90\%) \cite{Tri:16}. 

Following Rajpurkar et al. \shortcite{Raj:16} and Trischler et al. \shortcite{Hil:15}, we categorize answers based on their linguistic types such as time (N1), other numeric (N2), person (E1), location (E2), other entity (E3), noun phrase (P1), verb phrase (P3), adjective phrase (P3), preposition phrase (P4), clause (P5) and others (O). Unlike SQuAD \cite{Raj:16} and NewsQA \cite{Hil:15}, instead of using automatic tools for annotation, the answer types on the 
Dev set of UIT-ViQuAD are annotated entirely by workers. Table \ref{tab:answertype} shows the distribution of the answer types based on various syntactic structures on the Dev set of our dataset. Common noun phrases account for the largest proportion in UIT-ViQuAD, which is similar to the statistics of SQuAD \cite{Raj:16} and NewsQA \cite{Tri:16}. In addition, verb phrases (P2) and other entities (E3) rank the second and third percentages in our dataset.
\begin{table}[H]
\centering

\begin{tabular}{cccccccccccc}
\hline
{\bf Answer type} & {\bf N1} & {\bf N2} & {\bf E1} & {\bf E2} & {\bf E3} & {\bf P1} & {\bf P2} & {\bf P3} & {\bf P4} & {\bf P5} & {\bf O} \\ \hline
{Percentage}     &7.71   &9.41   &5.39   &4.32   &{\bf 11.65}   &{\bf 22.86}   &{\bf 18.43}   &2.52   &3.18   &5.91    &10.55    \\    \hline
                           
\end{tabular}
\caption{Statistics of the answer types on the Dev set of the UIT-ViQuAD dataset.}
\label{tab:answertype}
\end{table}

\section{Empirical evaluation}
In this section, we conduct experiments with the state-of-the-art MRC models to evaluate our dataset. To measure the difficulty of our dataset, we also estimate human performance on the task of Vietnamese MRC. Similar to evaluations on English and Chinese datasets \cite{Raj:16,Cui:18}, we used two evaluation metrics, exact match (EM) and F1-score, to evaluate performances of MRC models on our dataset.

\subsection{Human performance}

In order to measure human performance on the development and test sets, we hired three other workers to independently answer questions on the test and development sets. As a result, each question in the development ans test sets has four answers, as described in Phase 5 of Section \ref{datacreation}. Unlike Rajpurkar et al. \shortcite{Raj:16} and like Cui et al. \shortcite{Cui:18},  to measure the performance, we use a cross-validation methodology. In particular, we consider the first answer as human prediction and treat the remainder of the answers as ground truths. We obtain three human prediction performances by iteratively regarding the first, second, and third answer as the human prediction. We take the maximum performance over all of the ground truth answers for each question. Lastly, we calculate the average of four results as the final human performance on the dataset.

\subsection{Re-implemented methods and baselines}
In this paper, we re-implemented the following MRC models on our dataset as described in Section \ref{datasetanalysis}.
\begin{itemize}
    \item \textbf{DrQA}: Chen et al. \shortcite{Che:2017} introduced a simple but effective neural network-based model for the MRC task. DrQA Reader achieved good performance on multiple MRC datasets \cite{Raj:16,Red:19,Lab:19}. Thus, we re-implement this method into our dataset as the first baseline models to compare future models.
    
    \item \textbf{QANet}: QANet was proposed by Yu et al. \shortcite{Yu:18} and this model also demonstrated good performance on multiple MRC datasets \cite{Raj:16,Dua:19}. This model consists of multiple convolutional layers followed by two components: the self-attention and fully connected layer, for both question and passage encoding as well as some more layers stacked before predicting the final output.

    \item \textbf{BERT}: BERT was proposed by Devlin et al. \shortcite{Dev:18}. This model is a strong methodology for pre-training language representations, which achieved the state-of-the-art results on many reading comprehension tasks. In this paper, we used mBERT \cite{Dev:18}, a large-scale multilingual language model pre-trained for the evaluation of our Vietnamese MRC task.
    
    \item \textbf{XLM-R}: XLM-R was proposed by Conneau et al. \shortcite{Con:19}, a super strong methodology for pre-training multilingual language models at scale, which leads to significant performance gains for a wide range of cross-lingual transfer tasks. This model significantly outperforms multilingual BERT (mBERT) on a variety of crosslingual benchmarks, including XNLI, MLQA, and NER. In this paper, we evaluate XLM-R$_{\text{Base}}$ and XLM-R$_\text{Large}$ on our dataset.

\end{itemize}

\subsection{Experimental settings}
We use a single NVIDIA Tesla P100 GPU via Google Colaboratory to train all MRC models on our dataset. We utilize the pre-trained word embeddings introduced by \cite{vu2019etnlp}, including Word2vec, fastText, ELMO, and BERT$_{Base}$ for DrQA and QANet. Besides, we set \textit{batch size = 32} and \textit{epochs = 40} for both the two models. To evaluate BERT on our dataset, we implement a multilingual pre-trained model mBERT \cite{Dev:18} and pre-trained cross-lingual models XLM-R \cite{Con:19} with the baseline configuration provided by HuggingFace\footnote{https://huggingface.co}. Based on our dataset characteristics, we use the maximum answer length to 300, the question length to 64, and the input sequence length to 384 for all the experiments on mBERT and XLM-R.

\begin{table}[H]
\centering
\begin{tabular}{lrrrr}
\hline
\multicolumn{1}{c}{\multirow{2}{*}{\textbf{Model}}} & \multicolumn{2}{c}{\textbf{EM}}                                       & \multicolumn{2}{c}{\textbf{F1-score}}                                 \\ \cline{2-5} 
\multicolumn{1}{c}{}                                & \multicolumn{1}{c}{\textbf{Dev}} & \multicolumn{1}{c}{\textbf{Test}} & \multicolumn{1}{c}{\textbf{Dev}} & \multicolumn{1}{c}{\textbf{Test}} \\ \hline

DrQA + Word2vec &39.04        &38.10      &60.31      &60.38                                \\
DrQA + FastText  &35.93     &35.61     &59.33     &58.67                              \\

DrQA + ELMO &43.98       &{\bf 40.91}     &65.09     &{\bf 63.44}                               \\

DrQA + BERT   &35.71       &34.84     &58.00     &57.73                             \\ \hline

QANet + Word2vec  &45.19     &40.89      &67.73       &64.60                              \\ 
QANet + FastText &39.66       &{\bf 46.05}     &63.82     &{\bf 68.06}                               \\
QANet + ELMO  &46.10       &42.21     &67.62     &65.76                                  \\ 
QANet + BERT  & 43.13   & 41.93     & 66.54     & 65.45                              \\ \hline
mBERT   & 62.20                             & {\bf 59.28}                              & 80.77                             & {\bf 80.00}                               \\ \hline
XLM-R$_{\text{Base}}$                                          & 63.87                             & 63.00                              & 81.90                             & 81.95                               \\ 
XLM-R$_{\text{Large}}$                                         & {\bf 69.18}                             & {\bf 68.98}                              & {\bf 87.14}                             & {\bf 87.02}                              \\ \hline
Human performance                                                 & 85.65                             & 85.59                              & 95.19                             & 94.69                               \\ \hline
\end{tabular}
\caption{Human and model performances on the Dev and Test sets of UIT-ViQuAD.}\label{tab:performace}
\end{table}

\subsection{Evaluation results}
Table \ref{tab:performace} presents the performance of our models alongside human performance on the development and test sets of our dataset. For EM and F1-core, XLM-R$_{\text{Large}}$ significantly outperforms the other models but is largely below human performance. On the test set, the model predicts answers with the F1-score of 87.02\%. However, this model's exact match achieves 68.98\%, which is significantly lower than the F1-score.

\subsection{Analysis}
To gain more in-depth insights into the evaluation of the machine models and humans in Vietnamese, we analyze their performances in terms of different linguistic aspects such as length-based (question length, answer length, and passage length) and type-based (question type, answer type, and reasoning type).

\subsubsection{Effects of length-based aspects}
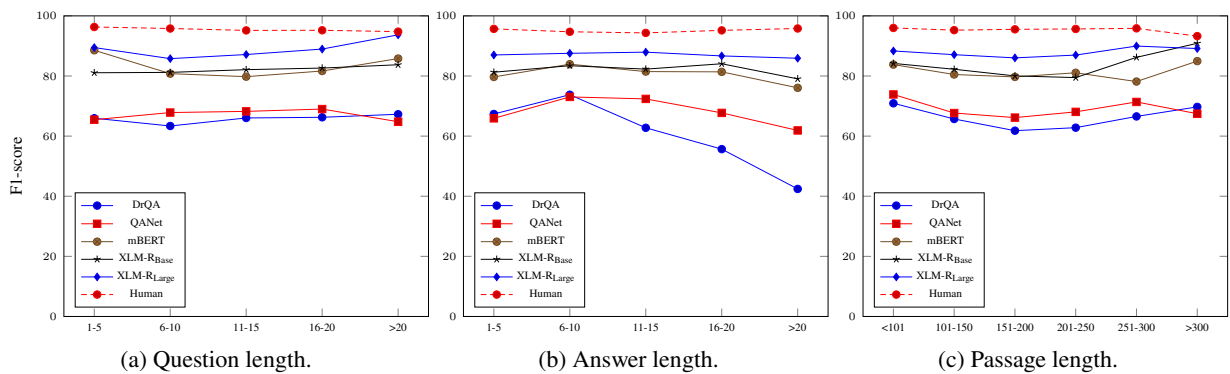
\begin{figure}[H]
    \centering
    \hspace{-1cm}
    \subfloat[Question length.]{\begin{tikzpicture}[scale = 0.7]
        \begin{axis}[
            legend entries={DrQA, QANet,mBERT,XLM-R$_{\text{Base}}$, XLM-R$_{\text{Large}}$, Human},
            legend pos=south west,
            legend style={font=\tiny},
            xlabel near ticks,
            ylabel near ticks,
            xlabel style={font=\footnotesize},
            ylabel style={font=\footnotesize},
            ylabel=F1-score,
            xticklabels={{},{},{1-5},{6-10},{11-15},{16-20},{\textgreater20}},
            ymin=0,
            ymax=100,
            xticklabel style={
                font=\tiny,
                scaled x ticks = false,
            },
            yticklabel style={
                font=\tiny,
            },
            table/col sep=comma,
        ]
            \addplot table [x=t, y=xlmrl_f1] {questionlenth.csv};
            \addplot table [x=t, y=xlmrs_f1]  {questionlenth.csv};
            \addplot table [x=t, y=bert_f1]  {questionlenth.csv};
            \addplot table [x=t, y=qanet_f1]  {questionlenth.csv};
            \addplot table [x=t, y=drqa_f1]  {questionlenth.csv};
            \addplot table [x=t, y=human_f1]  {questionlenth.csv};
        \end{axis}
        \label{fig:questionlengthanalysis}
    \end{tikzpicture}}%
    \subfloat[Answer length.]{\begin{tikzpicture}[scale = 0.7]
        \begin{axis}[
            legend entries={DrQA, QANet,mBERT,XLM-R$_{\text{Base}}$, XLM-R$_{\text{Large}}$, Human},
            legend pos=south west,
            legend style={font=\tiny},
            xlabel near ticks,
            ylabel near ticks,
            xlabel style={font=\footnotesize},
            ylabel style={font=\footnotesize},
            xticklabels={{},{},{1-5},{6-10},{11-15},{16-20},{\textgreater20}},
            ymin=0,
            ymax=100,
            xticklabel style={
                /pgf/number format/set thousands separator={},
                font=\tiny,
                scaled x ticks = false,
            },
            yticklabel style={
                font=\tiny,
            },
            table/col sep=comma,
        ]
            \addplot table [x=t, y=xlmrl_f1] {answerlenth.csv};
            \addplot table [x=t, y=xlmrs_f1]  {answerlenth.csv};
            \addplot table [x=t, y=bert_f1]  {answerlenth.csv};
            \addplot table [x=t, y=qanet_f1]  {answerlenth.csv};
            \addplot table [x=t, y=drqa_f1]  {answerlenth.csv};
            \addplot table [x=t, y=human_f1]  {answerlenth.csv};
        \end{axis}
        \label{fig:answerlengthanalysis}
    \end{tikzpicture}}%
    \subfloat[Passage length.]{\begin{tikzpicture}[scale = 0.7]
        \begin{axis}[
            legend entries={DrQA, QANet,mBERT,XLM-R$_{\text{Base}}$, XLM-R$_{\text{Large}}$, Human},
            legend pos=south west,
            legend style={font=\tiny},
            xlabel near ticks,
            ylabel near ticks,
            xlabel style={font=\footnotesize},
            ylabel style={font=\footnotesize},
            xticklabels={{},{},{\textless101},{101-150},{151-200},{201-250},{251-300},{\textgreater300}},
            ymin=0,
            ymax=100,
            xticklabel style={
                /pgf/number format/set thousands separator={},
                font=\tiny,
                scaled x ticks = false,
            },
            yticklabel style={
                font=\tiny,
            },
            table/col sep=comma,
        ]
            \addplot table [x=t, y=xlmrl_f1] {passagelenth.csv};
            \addplot table [x=t, y=xlmrs_f1]  {passagelenth.csv};
            \addplot table [x=t, y=bert_f1]  {passagelenth.csv};
            \addplot table [x=t, y=qanet_f1]  {passagelenth.csv};
            \addplot table [x=t, y=drqa_f1]  {passagelenth.csv};
            \addplot table [x=t, y=human_f1]  {passagelenth.csv};
        \end{axis}
        \label{fig:passagelengthanalysis}
    \end{tikzpicture}}%
    \caption{Length-based analysis of F1-score performances on the Dev set of UIT-ViQuAD.}%
    \label{fig:lengthbasedanalysis}%
\end{figure}

In order to examine how well the MRC models could perform on UIT-ViQuAD, we analyze the performances of the machine models and humans by F1-score. Figure \ref{fig:lengthbasedanalysis} shows length-based analyses of humans and MRC models’ performances on the Dev set. In general, the performances of the mBERT and XLM-R models outperform that of the QANet and DrQA models. However, all machine models' performances are lower than humans on different types of lengths. For the question-length-based analysis (see Figure \ref{fig:questionlengthanalysis}), we found that longer questions tend to achieve better results because these questions maybe contain more information, which makes it easier for MRC models to find answers. On the contrary, the longer answers achieve lower performances, which is challenging for the MRC models, shown clearly in the performances of the DrQA and QANet models in Figure \ref{fig:answerlengthanalysis}. Unlike question-length and answer-length analyses, the passage lengths witness fluctuations in the performances of most MRC models work well for short (<100 words) and long (>250 words) passages (see Figure \ref{fig:passagelengthanalysis}). The result analyses based on the different lengths can be used to evaluate the difficulty of Vietnamese automatic reading comprehension on our dataset, which can help researchers have ideas for curriculum learning in future work.

\begin{figure}[H]
    \centering
    \subfloat[Question type.]{
\begin{tikzpicture}[scale=0.7]
    \begin{axis}[
        width  = 0.7*\textwidth,
        height = 10cm,
        major x tick style = transparent,
        ybar=1*\pgflinewidth,
        bar width=5pt,
        ymajorgrids = true,
        ylabel = {Percentage},
        enlarge x limits=0.1,
        symbolic x coords={Who,What,When,Where,Why,How,Others},
        xtick = data,
        ymin=0,
        scaled y ticks = false,
        legend cell align=left,
        legend style={
                at={(1,1.05)},
                anchor=south east,
                column sep=1ex
        }
    ]
    
        \addplot[pattern=dots]
             coordinates {(Who,65.37)(What,62.78)(When,81.59)(Where,60.75)(Why,63.54)(How,57.45)(Others,74.75)};
             
        \addplot[pattern=north east lines]
            coordinates {(Who,65.82)(What,66.5)(When,83.79)(Where,61.93)(Why,63.77)(How,66.44)(Others,73.7)};

        \addplot[pattern=grid]
             coordinates {(Who,83.61)(What,80.04)(When,87.81)(Where,74.05)(Why,77.52)(How,77.96)(Others,83.65)};
             
        \addplot[pattern=horizontal lines]
             coordinates {(Who,81.66)(What,87.25)(When,75.38)(Where,79.23)(Why,82.13)(How,79.4)(Others,84.47)};

        \addplot[style={black,fill=gray,mark=none}]
             coordinates {(Who,86.36)(What,88.4)(When,90.3)(Where,80.5)(Why,83.28)(How,82.75)(Others,89.5)};

        \addplot[style={style={black,fill=black,mark=none}}]
             coordinates {(Who,96.11)(What,95.83)(When,97)(Where,94.99)(Why,95.03)(How,94.22)(Others,93.72)};
             
        \addplot[dashed,smooth,mark=*] table[x index=0,y index=1,col sep=comma] {questype_average.csv};

    \end{axis}
\end{tikzpicture}}%
    \qquad
    \subfloat[Reasoning type.]{\begin{tikzpicture}[scale=0.7]
    \begin{axis}[
        width  = 0.54*\textwidth,
        height = 10cm,
        major x tick style = transparent,
        ybar=1*\pgflinewidth,
        bar width=5pt,
        ymajorgrids = true,
        ylabel = {Percentage},
        enlarge x limits=0.1,
        symbolic x coords={WM,PP,SSI,MSI,AoI},
        xtick = data,
        ymin=0,
        scaled y ticks = false,
        legend cell align=left,
        legend style={
                at={(1,1.05)},
                anchor=south east,
                column sep=1ex
        }
    ]
    
        \addplot[pattern=dots]
             coordinates {(WM,81.69)(PP,71.63)(SSI,57.97)(MSI,55.68)(AoI,55.77)};
             
        \addplot[pattern=north east lines]
            coordinates {(WM,83.93)(PP,73.96)(SSI,62.57)(MSI,56.16)(AoI,61.67)};

        \addplot[pattern=grid]
             coordinates {(WM,91.99)(PP,83.83)(SSI,76.5)(MSI,74.96)(AoI,76.69)};
             
        \addplot[pattern=horizontal lines]
             coordinates {(WM,93.12)(PP,86.26)(SSI,76.85)(MSI,76.29)(AoI,72.39)};

        \addplot[style={black,fill=gray,mark=none}]
             coordinates {(WM,95.23)(PP,90)(SSI,83.83)(MSI,83.24)(AoI,74.33)};

        \addplot[style={style={black,fill=black ,mark=none}}]
             coordinates {(WM,96.83)(PP,96.38)(SSI,95.12)(MSI,93.84)(AoI,90.07)};
             
        \addplot[dashed,smooth,mark=*] table[x index=0,y index=1,col sep=comma] {anwtype_average.csv};

    \end{axis}
\end{tikzpicture}}%
    \qquad
    \subfloat[Answer type.]{\begin{tikzpicture}[scale=0.7]
    \begin{axis}[
        legend entries={DrQA, QANet,mBERT,XLM-R$_{\text{Base}}$, XLM-R$_{\text{Large}}$, Human, F1-score Average},
        legend pos=outer north east,
        legend cell align=right,
        width  = 1.05*\textwidth,
        height = 10cm,
        ybar=1*\pgflinewidth,
        bar width=5pt,
        ymajorgrids = true,
        ylabel = {Percentage},
        enlarge x limits=0.05,
        symbolic x coords={N1,N2,E1,E2,E3,P1,P2,P3,P4,P5,O},
        xtick = data,
        ymin=0,
        scaled y ticks = false,
    ]
    
        \addplot[pattern=dots]
             coordinates {(N1,82.8)(N2,74.01)(E1,72)(E2,63.17)(E3,69.34)(P1,59.36)(P2,62.3)(P3,62.12)(P4,77.54)(P5,65.51)(O,50.2)};
             
        \addplot[pattern=north east lines]
            coordinates {(N1,86.71)(N2,73.35)(E1,72.06)(E2,64.74)(E3,70.3)(P1,60.56)(P2,69.66)(P3,50.29)(P4,78.08)(P5,69.8)(O,58.9)};

        \addplot[pattern=grid]
             coordinates {(N1,89.12)(N2,82.49)(E1,85.53)(E2,74.4)(E3,81.28)(P1,79.61)(P2,80.79)(P3,74.87)(P4,86.3)(P5,82.71)(O,71.9)};

        \addplot[pattern=horizontal lines]
             coordinates {(N1,88.69)(N2,82.87)(E1,86.39)(E2,76.69)(E3,82.43)(P1,79.53)(P2,81.78)(P3,82.81)(P4,81.58)(P5,84.32)(O,78.37)};

        \addplot[style={black,fill=gray,mark=none}]
             coordinates {(N1,90.9)(N2,88.87)(E1,90.69)(E2,81.96)(E3,89.4)(P1,85.46)(P2,87.27)(P3,86.96)(P4,89.48)(P5,88.4)(O,81.07)};

        \addplot[style={style={black,fill=black ,mark=none}}]
             coordinates {(N1,97.22)(N2,93.52)(E1,97.4)(E2,96.25)(E3,96.04)(P1,95.54)(P2,95.02)(P3,93.13)(P4,95.46)(P5,96.14)(O,94.16)};
             
        \addplot[dashed,smooth,mark=*] table[x index=0,y index=1,col sep=comma] {restype_average.csv};

        
    \end{axis}
\end{tikzpicture}}%
    \caption{Type-based analysis of F1-score performances on the Dev set of UIT-ViQuAD. The lines on the graphs are the average of F1-score performances on the MRC models.}%
    \label{fig:typebasedanalysis}%
\end{figure}
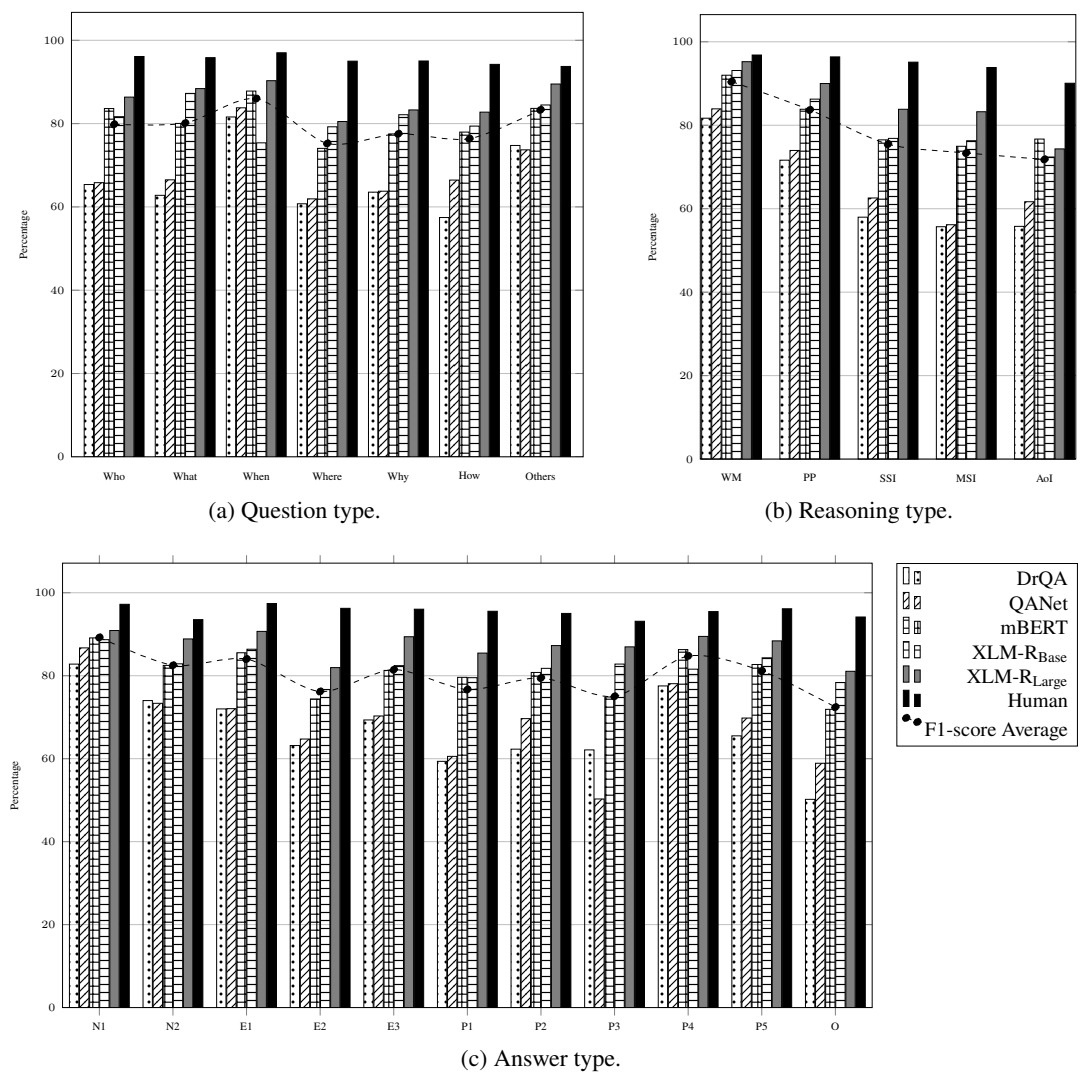

\subsubsection{Effects of type-based aspects} 
Besides, we examine how MRC models solve the type-based aspects of UIT-ViQuAD. Therefore, we analyze the F1-score performances of the machine models and humans on the development. Figure \ref{fig:typebasedanalysis} shows the type-based analyses of humans and MRC models' performances. No machine models have been able to handle question types, answer types, and types of reasoning better than humans. On the type of reasoning, complex inference types (SSI, MSI, and AoI) obtain lower performances, which is similar to results on SQuAD and NewsQA \cite{Tri:16}. Similarly, difficult question types (Why and How) obtain low performances. However, the Where question is also another question type that does not been handle well in machine models. Thus, the Location answer type related to the Where question type also achieves low performances. Although the noun-phrase answer type accounts for the highest proportion of the dataset (22.86\%), the machine model does not yet handle well as other types because of the diverse and complicated structure of Vietnamese noun phrases \cite{nguyen2018ensuring}.

\subsubsection{Effects of the amount of training data}

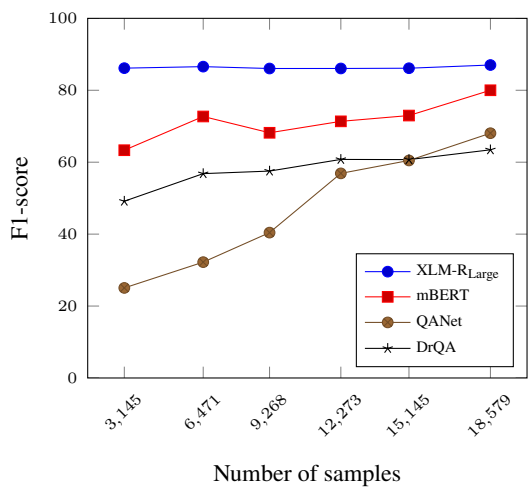
\begin{figure}[H]
\centering
\hspace{-1.4cm}
    \begin{tikzpicture}
        \begin{axis}[
            height=6.35cm,
            legend entries={XLM-R$_{\text{Large}}$,mBERT,QANet,DrQA},
            legend pos=south east,
            legend style={font=\tiny},
            legend cell align={left},
            xlabel near ticks,
            ylabel near ticks,
            xlabel style={font=\footnotesize},
            ylabel style={font=\footnotesize},
            xlabel=Number of samples,
            ylabel=F1-score,
            xtick={3145,6471,9268,12273,15145,18579},
            ymin=0,
            ymax=100,
            xticklabel style={
                rotate=45,
                font=\tiny,
                scaled x ticks = false,
            },
            yticklabel style={
                font=\tiny,
            },
            table/col sep=comma,
        ]
            \addplot table [x=t, y=xlmrl_f1] {results.csv};
            \addplot table [x=t, y=bert_f1]  {results.csv};
            \addplot table [x=t, y=qanet_f1]  {results.csv};
            \addplot table [x=t, y=drqa_f1]  {results.csv};
        \end{axis}
    \end{tikzpicture}
\hspace{-1cm}
   \caption{The impact of the amount of training data on the Test set of UIT-ViQuAD.}
   \label{fig:SizeDataToResult}
\end{figure}

The training data consists of 18,579 question-answer pairs which are lower than the quantity of the data trained for English and Chinese MRC models. To verify whether the small amount of training data affect the poor performance of the MRC systems based on model evaluations, we conduct various experiments with training sets comprising 3,145, 6,471, 9,268, 12,273, 15,145, and 18,579 questions. Figure \ref{fig:SizeDataToResult} shows the performance (F1-score) based on the Test set of UIT-ViQuAD. Through these experimental analyses, we find that DrQA, QANet, and mBERT obtain better performances when the amount of training data increases, whereas the performances of XLM-R are stable over 86\% with any training data amount. These observations indicate that the best model (XLM-R$_{\text{Large}}$) is more effective with a small amount of training data compared with the other three models. In general, increasing the training data quantity may be required to improve the performance of future models for most of neural network-based MRC models.

\section{Conclusion and future work}
In this paper, we introduce a new span-extraction dataset for evaluating Vietnamese MRC. UIT-ViQuAD contains over 23,000 questions generated by humans. Our experimental results show that the machines could obtain up to 87 percent scores on both the development and test set. However, they are lower than the estimated human performances in F1-score. We hope the release of our dataset contributes to the language diversity in MRC task, and accelerates further investigation on solving difficult questions that need comprehensive reasoning over multiple clues. According to the analysis results, we may extend this work by exploring models to solve challenging questions involving specific question types (Where, Why, and How), answer types (Location, and Noun Phrases) and reasoning types (Single-Sentence Inference, Multiple-Sentence Inference, Ambiguous or Insufficient). In future, we plan to enhance the quantity and the quality of our dataset to achieve better performance on deep learning and transformer models. In addition, we would like to open the Vietnamese MRC challenging task for researchers in the field.

\section*{Acknowledgements}
We would like to thank the reviewers' comments which are helpful for improving the quality of our work. In addition, we would like to thank our workers for their cooperation.

\bibliographystyle{coling}
\bibliography{coling2020}

\end{document}